\documentclass{INTERSPEECH2023}

\interspeechcameraready

\title{Improving Joint Speech-Text Representations Without Alignment}
\name{Cal Peyser$^{12}$, Zhong Meng$^2$, Ke Hu$^2$, Rohit Prabhavalkar$^2$, Andrew Rosenberg$^2$, Tara N. Sainath$^2$, Michael Picheny$^1$, Kyunghyun Cho$^1$}
\address{
    $^1$Center for Data Science, New York University, New York City, USA \\
    $^2$Google Inc., U.S.A}
\email{cpeyser@google.com}

\begin{document}

\maketitle
 
\begin{abstract}
The last year has seen astonishing progress in text-prompted image generation premised on the idea of a cross-modal representation space in which the text and image domains are represented jointly.  In ASR, this idea has found application as joint speech-text encoders that can scale to the capacities of very large parameter models by being trained on both unpaired speech and text.  While these methods show promise, they have required special treatment of the sequence-length mismatch inherent in speech and text, either by up-sampling heuristics or an explicit alignment model.  In this work, we offer evidence that joint speech-text encoders naturally achieve consistent representations across modalities by disregarding sequence length, and argue that consistency losses could forgive length differences and simply assume the best alignment.  We show that such a loss improves downstream WER in both a large-parameter monolingual and multilingual system.

\end{abstract}

\section{Introduction \label{sec:introduction}}
The power of very large models trained on vast unsupervised corpora in a single modality has become increasingly clear.  This has been demonstrated in the text domain where language models have achieved unprecedented zero-shot capabilities \cite{Brown20, Chowdhery22}, as well as in the audio domain, in which a single model has been shown to be adaptable to a surprisingly wide array of acoustic tasks \cite{Yang21, Borsos22}.  These successes have given rise to the question of how to apply these methods for problems involving two modalities, which historically have depended on manually paired data.

One very promising solution to this problem is to train a large encoder on both modalities, such that either modality may be provided as an unpaired example, but which learns to map paired examples to similar points in representation space.  In the image/text domain, such a representation has proved achievable and capable of attaining state-of-the-art performance on many image and text comprehension tasks in a single model \cite{Alayrac22, Cho21}.

In the audio/text domain, joint speech and text models have been utilized for a wide range of tasks \cite{Renduchintala18, Huang20, Mariooryad22}.  In speech recognition, the past few years has seen a trend toward models with a joint speech and text encoder to allow pretraining on unpaired speech and text data \cite{Tang20, Bapna22, Chen22, Sainath22}.  However, speech recognition presents the particular challenge of two sequence modalities, one of which (speech) is typically represented by a much longer sequence than the other (text).  This complicates the task of representing both modalities in the same embedding space, since we cannot make a direct, frame-wise comparison of an encoder's speech representation to its text representation. 

This complication has largely been handled either by upsampling or an explicit alignment model.  Fixed upsampling of the text inputs has been applied successfully for ASR in \cite{Sainath22} and SLU in \cite{Thomas22}, proving that an approximate alignment is sufficient for learning a joint representation.  On the other hand, \cite{Chen22_2} addresses the problem with a separately-trained alignment model that aims for perfect alignment.  In \cite{Chen22}, it's shown that such an alignment model permits the use of \say{consistency} regularization in which the encoder's outputs on corresponding speech and text are compared frame-wise and pushed together in representation space.  \cite{Chen22} goes on to show that \say{consistency} regularization yields a more closely joined representation space leading to better WER.   

Consistency regularization itself follows naturally from the literature on generative models.  Systems like autoencoders applied to augmented data (e.g. \cite{Chadebec_2022}) explicitly push representations of matched examples together, while contrastive systems like \cite{Chen20} do the same implicitly.  The success of the same idea in speech using an explicit alignment begs the question of if the same can be done with an implicit alignmentment; that is, without knowing the particular alignment between speech and text.

In this paper, we ask if consistency regularization may be applied using the implicit alignments learned in upsampling systems like \cite{Sainath22} to achieve the performance improvements seen with the explicit alignments in \cite{Chen22}.  To this end, we develop an algorithm inspired by dynamic time warping \cite{Sakoe78} that finds the \emph{best possible alignment} between an encoder's representation of a paired speech and text example.  We measure the quality of this \emph{best alignment} in a system without an explicit alignment model and show that that it is not only learned during training but in fact improves at deeper layers of the network.  Inspired by the improvements shown in \cite{Chen22_2} and \cite{Chen22}, we then show that by changing the criteria of the consistency regularization to encourage consistency under \emph{some} alignment, instead of a direct frame-wise comparison, we can achieve robust WER improvements against strong, semi-supervised baselines in both a monolingual and multilingual setting, all without any learned alignment model.  Our results suggest that enforcing consistency in cross-modal representations can be done by simply forgiving misalignment. 

The rest of this paper proceeds as follows.  Section 2 specifies our setup for joint speech/text modeling and consistency regularization, and the details of our best-alignment algorithm.  Section 3 specifies details of our data and training.  Section 4 presents our analysis of the best alignment in an unregularized model and the results of optimizing that alignment with our consistency loss.  We conclude in Section 5.

\section{Methods \label{sec:methods}}
In this section we present our setup for semi-supervised ASR by joint speech/text modeling, for which we mostly follow \cite{Sainath22}.  We then present our proposed best-alignment algorithm and define a corresponding consistency loss inspired by \cite{Chen22_2}.

\subsection{Model Architecture}
\begin{figure}[h]
  \centering
  \includegraphics[scale=0.4]{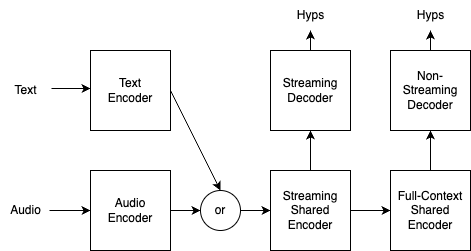}
  \caption{Our architecture for semi-supervised ASR, adapted from \cite{Sainath22} and \cite{Chen22}.}
   \label{fig:arch}
   \vspace{-0.1in}
\end{figure}

Figure \ref{fig:arch} gives our model architecture.  Essentially, we perform supervised ASR with streaming and non-streaming decoders, where the encoder is split into \say{audio-only}/\say{text-only} and \say{shared} components to permit text injection.  The simultaneous ASR and text-injection tasks give rise to a joint representation in the shared encoder.  Specifically, given a corpus of supervised examples $(x,y) \in \mathcal{S}$ and an unpaired text corpus $y \in \mathcal{U}$, our model contains the following components:

\begin{itemize}
  \item $E_a$: The audio encoder, which embeds audio features $x$.
  \item $E_t$: The text encoder, which embeds text features $y$.
  \item $E^{\text{stream}}_s$: The shared streaming encoder, which may consume either $E_a(x)$ or $E_t(y)$ and maps them to a joint representation.  Since this encoder is \say{streaming}, it only receives past acoustic frames.
  \item $E^{\text{full-context}}_s$: The full-context encoder, which consumes the outputs of $E^{\text{stream}}_s$ and which is given forward acoustic frames.
  \item $D^{\text{stream}}$: The streaming decoder, which consumes the outputs of $E^{\text{stream}}_s$ and emits streaming ASR hypothesis.
  \item $E^{\text{non-streaming}}_s$: The non-streaming decoder, which consumes the outputs of $E^{\text{full-context}}_s$ and emits non-streaming ASR hypothesis.
\end{itemize}

Our model is trained simultaneously on two tasks: ASR, and masked text reconstruction.  For ASR, audio is passed into audio encoder, and hypothesis are compared against ground truth text with the conventional cross-entropy loss.  Masked text reconstruction makes use of unpaired text data. A mask (15\% of the transcript) is applied to a phonemic representation of text, which is then passed into the text encoder.  The hypothesis is compared to the masked portion of the input again with a cross-entropy loss.

\subsection{Consistency Loss}
Consider a paired example $(x, y)$, where $x = (x_0,...,x_n)$ are speech inputs and $y = (y_0,...,y_m)$ are text inputs and where $n > m$.  Let us define the shared representations of audio and text as

\[R_a = E_s(E_a(x)) \quad R_t = E_s(E_t(y))\]

where $E_s$ can represent either the application of only $E^{\text{stream}}_s$ (for a streaming representation) of $E^{\text{stream}}_s$ followed by $E^{\text{non-streaming}}_s$ (for a non-streaming representation).  A \say{consistency loss}, as developed in \cite{Chen22_2} and \cite{Chen22}, is some loss $\mathcal{L}^{\text{consistency}}(R_a, R_t)$ that measures the similarity of the two representations.

Since the audio $x$ and the text $y$ are sequences of different lengths, we require some notion of an alignment to define a meaningful consistency loss.  By alignment, we mean a specific up-sampling of $y$ such that each audio frame $x[i]$ will correspond to some text frame $y[j]$.  With this in mind, we define an alignment $\mathcal{A} = (a_0, a_1,...,a_n)$ as a list of indexes into $y$, such that for any audio frame $i$, $x[i]$ corresponds to $y[a_i]$ in the alignment.  We will also add the constraint that $a_i \le a_{i+1}$ for all $i$.  That is, we constrain $\mathcal{A}$ to be monotonically increasing, so that so that sequential audio frames may not correspond to text backwards.

This formulation is one of many conceivable ways to define an \say{alignment} and we've chosen it for the practicality it offers in efficiently computing the best alignment (see Section \ref{sec:best_alignment} below).  We note that in this formulation, each audio frame is considered exactly once, while each text frame can be repeated or skipped over entirely.

With this definition in mind, we define the consistency loss for a given alignment as

\[ \mathcal{L}^{\text{consistency}}_{\mathcal{A}}(R_a, R_t) = \sum_{x=0}^n \frac{\mathcal{L}^{\text{frame}} (R_a[x], R_t[\mathcal{A}[x]])}{n} \]

where $\mathcal{L}^{\text{frame}}$ is some frame-wise similarity measure  (in this work, we use L2).  That is, $\mathcal{L}^{\text{consistency}}_{\mathcal{A}}(R_a, R_t)$ gives the average frame-wise similarity between $R_a$ and the specific up-sampling of $R_t$ given by $\mathcal{A}$.

The setups in \cite{Chen22_2} and \cite{Chen22} use such a consistency loss successfully, taking $\mathcal{A}$ from a neural alignment model.  We propose, as an alternative, to optimize the consistency over the \emph{best possible} alignment:

\[ \mathcal{L}^{\text{consistency}}(R_a, R_t) = \min_{\mathcal{A}}\mathcal{L}^{\text{consistency}}_{\mathcal{A}}(R_a, R_t)   \]

In order to train with such a loss, we require an efficient algorithm to compute the best alignment.

\subsection{Computing the Best Alignment} \label{sec:best_alignment}
Dynamic time warping \cite{Sakoe78} relies on an inductive rule in order to define a recursive algorithm to match two sequences based on a cost function.  We do the same, specifying the cost:

\[\mathcal{C}(i, j) = \min_{\mathcal{A}} \mathcal{L}^{\text{consistency}} (R_a[:i], R_t[:j]) \]

That is, the cost $\mathcal{C}(i, j)$ gives the consistency loss under the best alignment between the prefix of the audio representation up to the index $i$ and the prefix of the text representation up to the index $j$.  We may then specify a inductive rule:

\[\mathcal{C}(i, j) = \min_{k \le j} \left[ \mathcal{C}(i, k - 1) + \mathcal{L}^{\text{frame}}(R_a[i], R_t[k]) \right] \]

That is, the best alignment for the prefixes $R_a[:i]$ and $R_t[:j]$ aligns the previous $i-1$ audio frames to some shorter prefix $R_t[:k]$, and then appends to it the specific alignment of $R_a[i]$ to $R_t[k]$.

We may back out the indexes of the best alignment from this computation.  This rule gives rise to a dynamic programming algorithm for finding the best alignment in $\mathcal{O}(nm^2)$ time and memory.

We note that the minimization across all alignments precludes differentiation of the alignment-finding.  Instead, we compute the best alignment during forward-propagation, and then differentiate $\mathcal{L}^{\text{frame}}$ as applied to the aligned frames.  That is, we use the pass-through approximation of the gradient:

\[\frac{\partial \mathcal{L}^{\text{consistency}}(R_a, R_t)}{\partial \theta} \approx \frac{\partial \mathcal{L}^{\text{consistency}}_{\mathcal{A^*}}(R_a, R_t)}{\partial \theta} \]

where

\[\mathcal{A^*} = \text{arg}\min_{\mathcal{A}} \mathcal{L}^{\text{consistency}}_{\mathcal{A}}(R_a, R_t) \]

\section{Experiments \label{sec:experiments}}
In this section, we provide details of our model settings and data.

\subsection{Model Settings \label{sec:model_settings}}
We specify component's parameterizations according to the list in Section 2.1:

\begin{itemize}
  \item $E_a$: The audio encoder consists of a single conformer \cite{Gulati20} layer with 8 attention heads and dimension 2048.  The audio encoder consumes 128 dimensional log-mels spanning 32ms each and spaced apart by 10ms.  We then stack each frame with the frame before it and the two frames after it to yield 512 dimensional representations.  Finally, we subsample by taking each third frame, yielding a final frame rate of 30ms.
  \item $E_t$: The text encoder consists of a embedding projection followed by a conformer layer.  As in \cite{Sainath22}, we find it necessary to supply the text encoder with phonemic representations of text transcripts.  We then continue to follow \cite{Sainath22} by repeating each phoneme twice as an alignment heuristic. 
  \item $E^{\text{stream}}_s$: The shared streaming encoder consists of five conformer layers, with layer-norm applied at the end.
  \item $E^{\text{full-context}}_s$: The full-context shared encoder consists of nine additional conformer layers, with layer-norm applied at the end.
  \item $D^{\text{stream}}$: The streaming decoder is a HAT decoder \cite{Variani20} in which both the prediction and joint layers have dimension 640.
  \item $E^{\text{non-streaming}}_s$: The non-streaming decoder, is identical to the streaming decoder.
\end{itemize}

Together, our model contains about 165M parameters.  Training is done in two phases.  First, the audio encoder, joint encoders, and decoders are all trained on paired data for 800k steps with a batch size of 2048.  The text encoder is then added and the model is further trained with equally weighted supervised and unsupervised loss as described in Section 2.1, with the best alignment loss from Section 2.3 optionally included.  The model trains in this manner for 100k further steps with a batch size of 2048 for both the supervised and unsupervised data.

All models are implemented in Tensorflow, with the best alignment algorithm itself implemented as a CPU kernel.  We find that the addition of the best alignment computation does not significantly increase training time over the baseline model.

\subsection{Datasets \label{sec:datasets}}
Text-injection methods in ASR have historically been applied in two broad settings.  Strong baselines are often fine-tuned with very large text corpora to improve performance on difficult words.  Alternatively, text-injection may be used for models trained on limited supervised data may be used to improve the internal language model and get closer to a viable system.  With these two settings in mind, we study the best alignment loss in two setups: 

\begin{itemize}
    \item A large English corpus consisting of about 200k hours of supervised speech, together with an unsupervised text dataset of about 200B examples.
    
    We report results for a \textbf{Main} test set derived from the same distribution as the training examples, and a \textbf{Noisy} test set of especially noisy examples.
    \item A multilingual corpus consisting of the following eleven languages: English (\textbf{En}), French (\textbf{Fr}), Spanish (\textbf{Sp}), Arabic (\textbf{Ar}), Portuguese (\textbf{Po}), German (\textbf{De}), Russian (\textbf{Ru}), Hindi (\textbf{Hi}), Italian(\textbf{It}), Mandarin, and Japanese.  
    
    This setting involves no unsupervised text, with the MLM objective applied instead to the supervised transcripts.  The dataset consists of about 140M paired examples.
    
    Bolded abbreviations are given above for languages for which we are able to report WER in \ref{table:multilang_wer}.  For simplicity with the large number of test sets, we report only non-streaming WER  from this model.    
\end{itemize}

All datasets are anonymized and human transcribed.

\begin{table*}[t]
	\centering
	\begin{tabular}{|c|c|c|c|c|c|c|c|c|c|}
		\toprule
		 & \textbf{En} & \textbf{Fr} & \textbf{Sp} & \textbf{Ar} & \textbf{Po} & \textbf{De} & \textbf{Ru} & \textbf{Hi} & \textbf{It}  \\
		\midrule
		\textbf{M\_0} & 9.1 & 10.6 & 6.4 & 12.6 & 7.9 & 14.8 & 13.0 & 19.7 & 10.3 \\
		\textbf{M\_10} & \textbf{8.5} & 10.4 & \textbf{5.8} & \textbf{11.8} & \textbf{7.7} & \textbf{13.4} & \textbf{12.5} & 19.4 & \textbf{9.8} \\
		\textbf{M\_1} & \textbf{8.5} & 10.5 & 6.1 & 11.9 & 8.1 & 13.9 & 12.7 & \textbf{19.3} & 10.0 \\
		\textbf{M\_0.1} & 8.6 & \textbf{10.3} & 6.2 & 12.1 & 8.0 & 13.9 & 12.9 & 19.5 & 9.9 \\
		\textbf{M\_0.01} & 8.8 & 10.5 & 6.3 & 12.2 & 7.9 & 14.0 & 13.0 & 19.6 & 10.3 \\
		\bottomrule
	\end{tabular}
 	\vspace{0.1in}
	\caption{Evaluation Results for the Multilingual Setting.}
	\label{table:multilang_wer}
\end{table*}

\section{Results \label{sec:results}}
In this section, we seek to demonstrate that even without consistency regularization, our model learns an alignment between paired speech and text examples.  We then seek to show that optimizing this alignment with our proposed best-alignment consistency regularization improves WER.

\subsection{Best-Alignment in an Unregularized Model \label{sec:analysis}}
For this analysis, we use our baseline model from the monolingual setup as described in Section \ref{sec:model_settings}. Our objective is to measure $\mathcal{L}^{\text{consistency}}$ on a small set of random development examples for $R_s$ and $R_t$ taken at each of the first five conformer layers of the streaming joint encoder.  We interpret a lower value for $\mathcal{L}^{\text{consistency}}$ as reflecting a stronger implicit alignment between speech and text.   

For each layer $l$ we sample 2000 random pairs of audio and text embeddings and compute the mean $\mu_l$ and standard deviation $\sigma^2_l$ of the distribution of distances between pairs.  We then compare two alignments: the naive frame-wise alignment and our computed best alignment.  For each of these alignments $\mathcal{A}$, we report:

\[ \mu_l - \frac{\mathcal{L}^{\text{consistency}}_{\mathcal{A}}(R_s, R_t)}{\sigma^2_l}  \]

That is, we report the consistency in terms of the number standard deviations away from the mean, such that a result of 0 suggests that the alignment is no better than random and a result below 0 suggests that the alignment is stronger than random.

\begin{table}[h]
\centering
\caption{Consistency of the linear and best alignments at layers of the shared encoder.}
\vspace*{0.1in}
\begin{tabular}{||c|c|c||} 
 \hline
 Layer & Frame-wise Alignment & Best Alignment \\ [0.5ex] 
 \hline\hline
 \textbf{1} & -0.06 & -1.47 \\ 
 \hline
 \textbf{2} & -0.23 & -2.15 \\ 
 \hline
 \textbf{3} & -0.29 & -2.61 \\ 
 \hline
 \textbf{4} & -0.37 & -2.67 \\ 
 \hline
 \textbf{5} & -0.49 & -3.06 \\ 
 \hline
\end{tabular}
\centering
\label{table:consistency_table}
\end{table}

Table \ref{table:consistency_table} presents these measurements.  We see that while the consistency of the frame-wise alignment is close to that of the random alignment, the best alignment is considerably better than random.  Furthermore, the quality of the best alignment improves steadily as we progress deeper into the model.  That is, while text and speech are not modeled jointly at the frame level, there is \emph{some} alignment for which paired speech and text are mapped to similar points in the embedding space, and this alignment improves with the depth of the network.
\begin{figure}[h]
  \centering
  \begin{subfigure}[t]{0.8\columnwidth}
	  \includegraphics[width=1.0\columnwidth]{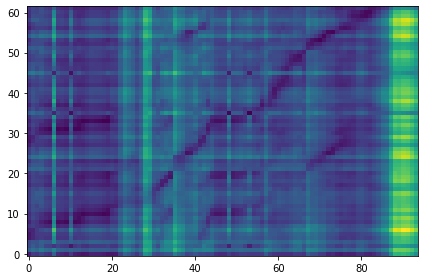}
	  \caption{\footnotesize Distances}
	  \label{fig:visualization_distances}
  \end{subfigure}
  \begin{subfigure}[t]{0.8\columnwidth}
    \includegraphics[width=1.0\columnwidth]{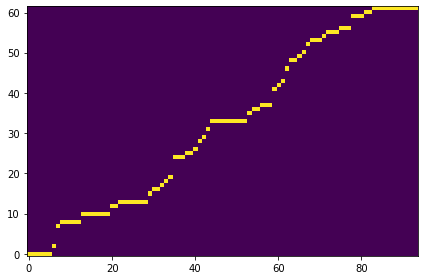}
	\caption{\footnotesize Best Alignment}
	\label{fig:visualization_best_alignment}
  \end{subfigure}
  \caption{Visualizations of embedding distances (a) and the best alignment (b) between an audio embedding on the horizontal axis and the corresponding text embedding on the vertical axis.  Darker points in (a) represent pairs of audio and text frames with nearby embeddings, and yellow points in (b) represent pairs in the recovered best alignment.}
  \label{fig:visualization}
\end{figure}

To illustrate the presence of this alignment, we visualize the relationship between shared encoder's final representations of the speech and text from a single test example.  Figure \ref{fig:visualization_distances} plots the distance between each pair of frames in the embeddings, and demonstrates that is indeed a single alignment with low distance.  Figure \ref{fig:visualization_best_alignment} shows how the best alignment algorithm recovers this trajectory. 

\begin{table}[h]
	\begin{subtable}[h]{0.2\textwidth}
    	\centering
    	\begin{tabular}{|c|c|c|c|}
    		\toprule
    		 & \textbf{Main} & \textbf{Noisy}  \\
    		\midrule
    		\textbf{E\_0} & 5.40 & 8.75 \\
    		\textbf{E\_10} & 5.37 & 8.70 \\
    		\textbf{E\_1} & 5.35 & \textbf{8.42} \\
    		\textbf{E\_0.1} & \textbf{5.27} & 8.77 \\
    		\textbf{E\_0.01} & 5.32 & 8.54\\
    		\bottomrule
    	\end{tabular}
     	\vspace{0.04in}
    	\caption{Non-Streaming}
	\end{subtable}
	\begin{subtable}[h]{0.3\textwidth}
	    \centering
    	\begin{tabular}{|c|c|c|c|}
    		\toprule
    		 & \textbf{Main} & \textbf{Noisy}  \\
    		\midrule
    		\textbf{E\_0} & 7.99 & 13.33 \\
    		\textbf{E\_10} & 8.21 & 13.08 \\
    		\textbf{E\_1} & 8.07 & 13.00 \\
    		\textbf{E\_0.1} & \textbf{7.90} & \textbf{12.63} \\
    		\textbf{E\_0.01} & 7.94 & 12.74 \\
    		\bottomrule
    	\end{tabular}
    	\vspace{0.1in}
    	\caption{Streaming}
	\end{subtable}
	\caption{Evaluation Results for the English-Only Setting.}
	\label{table:english_only_wer}
\end{table}

\subsection{Consistency Regularization Results \label{sec:bal_results}}
We present results for the best-alignment loss at different interpolation weights and for both of the settings specified in Section \ref{sec:datasets}.  For ease of reading, we specify each experiment by a letter and a number.  The letter is either $\mathbf{E}$ for the English-only setting of $\mathbf{M}$ for the multilingual setting.  The number is the interpolation weight of the best-alignment loss as a percentage.  For example, $\mathbf{E\_0}$ is the baseline English-only model with unregularized semisupervised finetuning, while $\mathbf{M\_0.1}$ is a multilingual model with the best alignment loss interpolated during finetuning at $0.1$ percent.

Table \ref{table:english_only_wer} gives our results in the high-resource, English-only setting.  There, we see small but consistent WER improvements with the best-alignment loss, although we note the necessity of selecting the correct interpolation weight.  Table \ref{table:multilang_wer} gives our results in the multilingual setting, where we see larger improvements.  We believe that the strength of the method in the multilingual setting is due to the increased difficulty of the problem and the smaller dataset leaving more room for the model to improve.

\section{Conclusions \label{sec:conclusions}}
We've shown that a semi-supervised speech/text encoder learns a joint representation of the two modalities that can observed by choosing the best alignment.  We've exploited that fact to enforce domain consistency with an extra loss term which optimizes the modality match for the best alignment.  We show consistent improvements over an unregularized joint model across multiple setups without adding any parameters.

\bibliographystyle{IEEEtran}
\bibliography{main}

\begin{thebibliography}{10}
\providecommand{\url}[1]{#1}
\csname url@samestyle\endcsname
\providecommand{\newblock}{\relax}
\providecommand{\bibinfo}[2]{#2}
\providecommand{\BIBentrySTDinterwordspacing}{\spaceskip=0pt\relax}
\providecommand{\BIBentryALTinterwordstretchfactor}{4}
\providecommand{\BIBentryALTinterwordspacing}{\spaceskip=\fontdimen2\font plus
\BIBentryALTinterwordstretchfactor\fontdimen3\font minus
  \fontdimen4\font\relax}
\providecommand{\BIBforeignlanguage}[2]{{%
\expandafter\ifx\csname l@#1\endcsname\relax
\typeout{** WARNING: IEEEtran.bst: No hyphenation pattern has been}%
\typeout{** loaded for the language `#1'. Using the pattern for}%
\typeout{** the default language instead.}%
\else
\language=\csname l@#1\endcsname
\fi
#2}}
\providecommand{\BIBdecl}{\relax}
\BIBdecl

\bibitem{Brown20}
T.~B. Brown, B.~Mann, N.~Ryder, M.~Subbiah, J.~Kaplan, P.~Dhariwal,
  A.~Neelakantan, P.~Shyam, G.~Sastry, A.~Askell, S.~Agarwal,
  A.~Herbert{-}Voss, G.~Krueger, T.~Henighan, R.~Child, A.~Ramesh, D.~M.
  Ziegler, J.~Wu, C.~Winter, C.~Hesse, M.~Chen, E.~Sigler, M.~Litwin, S.~Gray,
  B.~Chess, J.~Clark, C.~Berner, S.~McCandlish, A.~Radford, I.~Sutskever, and
  D.~Amodei, ``Language models are few-shot learners,'' in \emph{Advances in
  Neural Information Processing Systems}, 2020.

\bibitem{Chowdhery22}
A.~Chowdhery, S.~Narang, J.~Devlin, M.~Bosma, G.~Mishra, A.~Roberts, P.~Barham,
  H.~W. Chung, C.~Sutton, S.~Gehrmann, P.~Schuh, K.~Shi, S.~Tsvyashchenko,
  J.~Maynez, A.~Rao, P.~Barnes, Y.~Tay, N.~Shazeer, V.~Prabhakaran, E.~Reif,
  N.~Du, B.~Hutchinson, R.~Pope, J.~Bradbury, J.~Austin, M.~Isard, G.~Gur-Ari,
  P.~Yin, T.~Duke, A.~Levskaya, S.~Ghemawat, S.~Dev, H.~Michalewski, X.~Garcia,
  V.~Misra, K.~Robinson, L.~Fedus, D.~Zhou, D.~Ippolito, D.~Luan, H.~Lim,
  B.~Zoph, A.~Spiridonov, R.~Sepassi, D.~Dohan, S.~Agrawal, M.~Omernick, A.~M.
  Dai, T.~S. Pillai, M.~Pellat, A.~Lewkowycz, E.~Moreira, R.~Child, O.~Polozov,
  K.~Lee, Z.~Zhou, X.~Wang, B.~Saeta, M.~Diaz, O.~Firat, M.~Catasta, J.~Wei,
  K.~Meier-Hellstern, D.~Eck, J.~Dean, S.~Petrov, and N.~Fiedel, ``Palm:
  Scaling language modeling with pathways,'' 2022.

\bibitem{Yang21}
S.~Yang, P.~Chi, Y.~Chuang, C.~J. Lai, K.~Lakhotia, Y.~Y. Lin, A.~T. Liu,
  J.~Shi, X.~Chang, G.~Lin, T.~Huang, W.~Tseng, K.~Lee, D.~Liu, Z.~Huang,
  S.~Dong, S.~Li, S.~Watanabe, A.~Mohamed, and H.~Lee, ``{SUPERB:} speech
  processing universal performance benchmark,'' in \emph{INTERSPEECH}, 2021.

\bibitem{Borsos22}
Z.~Borsos, R.~Marinier, D.~Vincent, E.~Kharitonov, O.~Pietquin, M.~Sharifi,
  O.~Teboul, D.~Grangier, M.~Tagliasacchi, and N.~Zeghidour, ``Audiolm: a
  language modeling approach to audio generation,'' 2022.

\bibitem{Alayrac22}
J.-B. Alayrac, J.~Donahue, P.~Luc, A.~Miech, I.~Barr, Y.~Hasson, K.~Lenc,
  A.~Mensch, K.~Millican, M.~Reynolds, R.~Ring, E.~Rutherford, S.~Cabi, T.~Han,
  Z.~Gong, S.~Samangooei, M.~Monteiro, J.~Menick, S.~Borgeaud, A.~Brock,
  A.~Nematzadeh, S.~Sharifzadeh, M.~Binkowski, R.~Barreira, O.~Vinyals,
  A.~Zisserman, and K.~Simonyan, ``Flamingo: a visual language model for
  few-shot learning,'' in \emph{Advances in Neural Information Processing
  Systems}, 2022.

\bibitem{Cho21}
J.~Cho, J.~Lei, H.~Tan, and M.~Bansal, ``Unifying vision-and-language tasks via
  text generation,'' in \emph{International Conference on Machine Learning},
  2021.

\bibitem{Renduchintala18}
A.~Renduchintala, S.~Ding, M.~Wiesner, and S.~Watanabe, ``Multi-modal data
  augmentation for end-to-end {ASR},'' in \emph{INTERSPEECH}, 2018.

\bibitem{Huang20}
Y.~Huang, H.~Kuo, S.~Thomas, Z.~Kons, K.~Audhkhasi, B.~Kingsbury, R.~Hoory, and
  M.~Picheny, ``Leveraging unpaired text data for training end-to-end
  speech-to-intent systems,'' in \emph{IEEE International Conference on
  Acoustics, Speech and Signal Processing}, 2020.

\bibitem{Mariooryad22}
S.~Mariooryad, M.~Shannon, S.~Ma, T.~Bagby, D.~Kao, D.~Stanton, E.~Battenberg,
  and R.~Skerry-Ryan, ``Learning the joint distribution of two sequences using
  little or no paired data,'' 2022.

\bibitem{Tang20}
Y.~Tang, J.~M. Pino, C.~Wang, X.~Ma, and D.~Genzel, ``A general multi-task
  learning framework to leverage text data for speech to text tasks,'' in
  \emph{IEEE International Conference on Acoustics, Speech and Signal
  Processing}, 2021.

\bibitem{Bapna22}
A.~Bapna, C.~Cherry, Y.~Zhang, Y.~Jia, M.~Johnson, Y.~Cheng, S.~Khanuja,
  J.~Riesa, and A.~Conneau, ``mslam: Massively multilingual joint pre-training
  for speech and text,'' 2020.

\bibitem{Chen22}
Z.~Chen, Y.~Zhang, A.~Rosenberg, B.~Ramabhadran, P.~Moreno, A.~Bapna, and
  H.~Zen, ``Maestro: Matched speech text representations through modality
  matching,'' in \emph{INTERSPEECH}, 2022.

\bibitem{Sainath22}
T.~N. Sainath, R.~Prabhavalkar, A.~Bapna, Y.~Zhang, Z.~Huo, Z.~Chen, B.~Li,
  W.~Wang, and T.~Strohman, ``Joist: A joint speech and text streaming model
  for asr,'' in \emph{IEEE Spoken Language Technology Workshop}, 2022.

\bibitem{Thomas22}
S.~Thomas, H.-K.~J. Kuo, B.~Kingsbury, and G.~Saon, ``Towards reducing the need
  for speech training data to build spoken language understanding systems,'' in
  \emph{IEEE International Conference on Acoustics, Speech and Signal
  Processing}, 2022.

\bibitem{Chen22_2}
Z.~Chen, Y.~Zhang, A.~Rosenberg, B.~Ramabhadran, P.~Moreno, and G.~Wang,
  ``Tts4pretrain 2.0: Advancing the use of text and speech in asr pretraining
  with consistency and contrastive losses,'' in \emph{IEEE International
  Conference on Acoustics, Speech and Signal Processing}, 2022.

\bibitem{Chadebec_2022}
C.~Chadebec, E.~Thibeau-Sutre, N.~Burgos, and S.~Allassonniere, ``Data
  augmentation in high dimensional low sample size setting using a
  geometry-based variational autoencoder,'' in \emph{{IEEE} Transactions on
  Pattern Analysis and Machine Intelligence}, 2022.

\bibitem{Chen20}
T.~Chen, S.~Kornblith, M.~Norouzi, and G.~Hinton, ``A simple framework for
  contrastive learning of visual representations,'' in \emph{International
  Conference on Machine Learning}, 2020.

\bibitem{Sakoe78}
H.~Sakoe and S.~Chiba, ``Dynamic programming algorithm optimization for spoken
  word recognition,'' in \emph{IEEE Transactions on Acoustics, Speech, and
  Signal Processing}, 1978.

\bibitem{Gulati20}
A.~Gulati, J.~Qin, C.-C. Chiu, N.~Parmar, Y.~Zhang, J.~Yu, W.~Han, S.~Wang,
  Z.~Zhang, Y.~Wu, and R.~Pang, ``Conformer: Convolution-augmented transformer
  for speech recognition,'' in \emph{INTERSPEECH}, 2020.

\bibitem{Variani20}
E.~Variani, D.~Rybach, C.~Allauzen, and M.~Riley, ``Hybrid autoregressive
  transducer (hat),'' in \emph{IEEE International Conference on Acoustics,
  Speech and Signal Processing}, 2020.

\end{thebibliography}

\end{document}